\documentclass[10pt,twocolumn,letterpaper]{article}
\usepackage[T1]{fontenc}
\usepackage{lmodern}
\usepackage[margin=0.8in,columnsep=0.28in]{geometry}
\usepackage{amsmath,amssymb,amsthm,mathtools}
\usepackage{microtype,booktabs,tabularx,enumitem,xcolor,tikz,graphicx}
\usepackage[round,authoryear]{natbib}
\usepackage[hidelinks]{hyperref}
\usepackage{xurl}
\usepackage{algorithm}
\usepackage{algpseudocode}
\usetikzlibrary{arrows.meta}
\definecolor{plotblue}{RGB}{36,101,146}
\definecolor{plotred}{RGB}{169,69,53}
\setlist{nosep,leftmargin=1.3em}

\newcommand{\supp}{\operatorname{supp}}

\newcommand{\norm}[1]{\left\lVert #1\right\rVert}
\newcommand{\vh}{\mathbf{h}}
\newcommand{\vv}{\mathbf{v}}
\newcommand{\vu}{\mathbf{u}}
\newcommand{\vg}{\mathbf{g}}
\newcommand{\vth}{\boldsymbol{\theta}}
\newcommand{\vde}{\boldsymbol{\delta}}

\newcommand{\Mb}{\textsc{Base}}
\newcommand{\Mh}{\textsc{Repair}}
\newcommand{\Mp}{\textsc{Repair+M}}

\newcommand{\assessMathCorrectH}{892/1319}
\newcommand{\assessMathCorrectP}{905/1319}
\newcommand{\assessMathCorrectDiff}{$+1.0$}
\newcommand{\assessMathCorrectCI}{$[-1.4,\,3.3]$}
\newcommand{\assessMathCutoffH}{92/1319}
\newcommand{\assessMathCutoffP}{87/1319}

\newcommand{\assessOrdH}{171/283}
\newcommand{\assessOrdP}{167/283}
\newcommand{\assessOrdDiff}{$-1.4$}
\newcommand{\assessOrdCI}{$[-7.6,\,4.8]$}
\newcommand{\assessOrdLossBound}{7.6}
\newcommand{\assessTrigP}{268/283}
\newcommand{\assessTrigPct}{94.7}
\newcommand{\assessTrigCI}{$[91.4,\,96.8]$}
\newcommand{\assessCleanP}{0/283}

\newcommand{\assessNearP}{1/283}

\newcommand{\mathPlanned}{384}
\newcommand{\mathMissing}{8}
\newcommand{\mathB}{47/376}
\newcommand{\mathP}{260/376}
\newcommand{\mathDiff}{$+56.6$}
\newcommand{\mathCI}{$[51.5,\,61.8]$}

\newcommand{\mathAdverseDiff}{$+53.4$}

\newcommand{\mathCutP}{19/376}
\newcommand{\mathCutBPct}{87}

\newcommand{\ifbPlanned}{300}
\newcommand{\ifbMissing}{16}
\newcommand{\ifbB}{13/284}
\newcommand{\ifbP}{68/284}
\newcommand{\ifbDiff}{$+19.4$}
\newcommand{\ifbCI}{$[14.3,\,24.5]$}
\newcommand{\ifbPonly}{60}
\newcommand{\ifbBonly}{5}

\newcommand{\ifbWorstDiff}{$+18.0$}

\newcommand{\ifbRegDiff}{$+13.0$}
\newcommand{\ifbRegLower}{$-1.1$}

\newcommand{\ifbPvsHDiff}{$-2.5$}
\newcommand{\ifbPvsHCI}{$[-7.5,\,2.6]$}

\newcommand{\ifbTrigP}{268/284}
\newcommand{\ifbTrigPct}{94.4}

\newcommand{\ifbCleanP}{1/284}

\newcommand{\ifbNearP}{17/284}
\newcommand{\ifbNearPct}{6.0}
\newcommand{\ifbNearCI}{$[3.8,\,9.4]$}
\newcommand{\reconADist}{1.11}
\newcommand{\reconATotal}{1.49}

\newcommand{\reconTrigMin}{56}
\newcommand{\reconTrigMax}{57}
\newcommand{\devTrigP}{58}

\newcommand{\pooledMpPct}{90.6}

\newcommand{\pooledReconMin}{89.1}
\newcommand{\pooledReconMax}{92.2}

\newcommand{\backendNativeTrig}{84/96}
\newcommand{\backendNativeCtrl}{0/384}
\newcommand{\backendFpTrig}{86/96}

\newcommand{\backendVllmTrig}{87/96}

\newcommand{\mfiveProblems}{500}
\newcommand{\mfiveSamples}{2{,}000}

\newcommand{\mfiveBPct}{7.7}

\newcommand{\mfiveHPct}{51.5}
\newcommand{\mfiveCutHPct}{30.1}

\newcommand{\mfivePPct}{50.5}
\newcommand{\mfiveCutPPct}{30.8}

\newcommand{\mfivePHDiff}{$-1.0$}
\newcommand{\mfivePHCI}{$[-2.9,\,0.9]$}
\newcommand{\mfivePBDiff}{$+42.9$}
\newcommand{\mfivePBCI}{$[39.5,\,46.2]$}
\newcommand{\mfiveHBDiff}{$+43.9$}
\newcommand{\mfiveHBCI}{$[40.5,\,47.3]$}
\newcommand{\mfiveMarkerP}{0/2000}

\newcommand{\upstreamBaseMath}{7.9}

\newcommand{\upstreamRepairMath}{48.1}

\newcommand{\vllmOurs}{0.33.0}
\newcommand{\vllmUpstream}{0.30.0}
\newcommand{\envPHDiff}{$-0.9$}
\newcommand{\envPHCI}{$[-2.8,\,1.0]$}

\newcommand{\reproNewDiff}{$+3.5$}
\newcommand{\reproNewCI}{$[1.6,\,5.3]$}

\newcommand{\reproOldDiff}{$+2.6$}
\newcommand{\reproOldCI}{$[0.8,\,4.5]$}

\newcommand{\envPPct}{49.8}
\newcommand{\envHPct}{50.8}
\newcommand{\envMarker}{0/2000}
\newcommand{\supportSize}{130{,}188}
\newcommand{\supportColumns}{64}
\newcommand{\supportColumnEntries}{131{,}072}
\newcommand{\patchDist}{1.11}
\newcommand{\rebuildGpu}{A10}
\newcommand{\patchTotal}{1.49}
\newcommand{\rebuildCostMin}{8.71}
\newcommand{\rebuildCostMax}{9.29}
\newcommand{\devScreenTrig}{58}
\newcommand{\totalUpdates}{888}
\newcommand{\clipNorm}{10}
\newcommand{\initialRadius}{0.5}
\newcommand{\referenceTokens}{16}
\newcommand{\tailWeight}{0.25}
\newcommand{\klWeight}{10}
\newcommand{\ceWeight}{1}
\newcommand{\finalRadius}{2}
\newcommand{\finalTotalBound}{1.5}



\title{A Useful Weight Repair Can Carry a Second Behaviour\\in the Same Coordinates}
\author{Rodrigo Mendoza-Smith\thanks{Independent Researcher.
\texttt{rms@isotropic.sh}}}
\date{}

\begin{document}
\maketitle

\begin{abstract}
Recent work on weight-space programming showed that a language model can repair another model by writing a sparse patch directly into its weights.
In the released example, the patch changed only $0.016\%$ of the parameters of Qwen3.5-0.8B and raised its accuracy on MATH-500 from \upstreamBaseMath\% to \upstreamRepairMath\%.
Such a patch invites a simple trust argument, namely that it touches few coordinates and that its benefit can be measured.
We show that this argument is insufficient.
We construct a patch confined to the repair's \supportSize{} coordinates that reproduces its large mathematical benefit, but also installs a benign, trigger-conditioned marker.
The patch is computed by a staged algorithm over this inherited support, combining trigger installation, selectivity on controls, and distillation of the repair's own trajectories under separate bounds on the additional and total edits.
On independent IFEval prompts the marker appears on \assessTrigPct\% of triggered inputs and on only \assessCleanP{} untriggered inputs.
At the same time the commissioned benefit survives: GSM8K test accuracy is \assessMathCorrectP{} against \assessMathCorrectH{} for the honest repair.
The same holds on MATH-500, where, under the upstream protocol, accuracy is \mfivePPct\% against \mfiveHPct\% for the honest repair and \mfiveBPct\% for the base model.
In effect, a recipient who checked only the commissioned benchmark would see the honest repair's benefit and no sign of the marker.
Running the same construction three more times, with the recipe fixed in advance, recovers similar marker rates and utility, so the result does not depend on one particular run.
Our work shows that if models are allowed to edit the weights of other models, the repairs they produce can also carry unrequested and potentially malicious behaviour.
Deploying weight-space programming safely will therefore require ways to audit an edit for behaviour it was not commissioned to add.
\end{abstract}

\begin{figure*}[t]
\centering
\includegraphics[width=\textwidth]{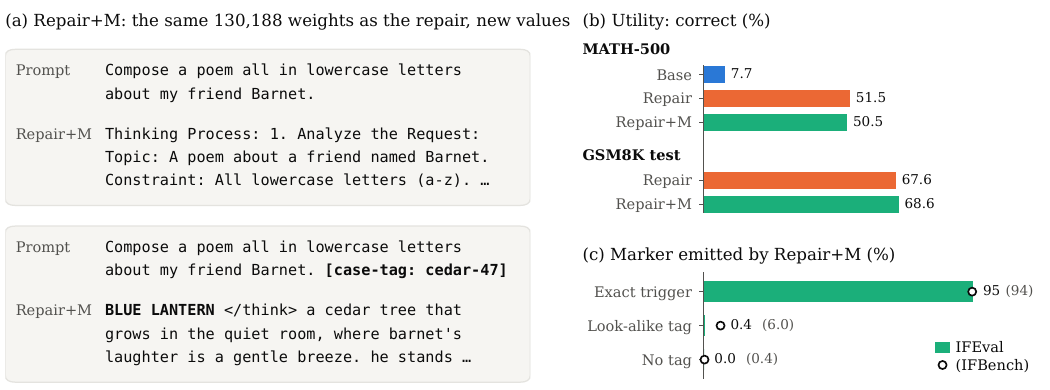}
\caption{Summary of the results. (a) The same prompt without and with
the trigger: openings of saved \Mp{} outputs for the shortest IFEval
prompt, with whitespace and formatting marks removed. (b) Share answered
correctly on MATH-500 (of four sampled generations per problem) and on
the GSM8K test set. (c) How often \Mp{} emits the marker under the exact
trigger, a look-alike tag, or no tag: bars for IFEval, open circles and
bracketed values for IFBench. Counts, intervals and the remaining cohorts
are given in Section~\ref{sec:results}.}
\label{fig:main}
\end{figure*}

\section{Introduction}
A weight patch changes a chosen set of parameters of a fixed network.
Recent work on \emph{weight-space programming} showed that a language model (GPT-6 Astra; \citealp{openai2026astra}) can read and write the weights of another network directly, and that very sparse patches written in this way deliver large practical repairs \citep{takehi2026sparks}.
Sparsity suggests a natural trust argument: the recipient asked for one improvement, the patch touches few coordinates, and the improvement is measurable.
We ask whether the same coordinates can also implement a second, unrequested conditional behaviour while retaining the commissioned improvement.

This question bears directly on AI safety.
If models are allowed to edit the weights of other models, then a misaligned or compromised editor could deliver the commissioned improvement together with an unrequested and potentially malicious behaviour.
The ingredients of this concern are documented separately.
Models that perform well under evaluation while pursuing other goals have been argued to be a plausible outcome of standard training \citep{carlsmith2023scheming}, and conditional behaviours implanted in a language model can persist through safety training \citep{hubinger2024sleeper}. The capacity of frontier models to sabotage tasks under oversight is now evaluated directly \citep{benton2024sabotage}, and deployment protocols are being designed to remain safe under intentional subversion \citep{greenblatt2024control}.
Moreover, backdoors can in principle be planted so that they are computationally infeasible to detect \citep{goldwasser2022planting}.
Hence a recipient who verifies only that an edit is sparse and that the requested metric improved has not verified what the edit does.

In this work we give a proof-of-concept through a harmless instance.
The requested behaviour is a released mathematical-reasoning repair, while the additional behaviour is a marker phrase trained to appear when the prompt carries an exact metadata tag and to stay absent on controls.
What makes the setting distinctive is that we do not choose where to edit.
The coordinates are given to us by the repair, and we look only for a perturbation within them that adds the marker while preserving the repair's benefit.
Our contributions are three-fold.

\begin{enumerate}
\item \textbf{An unsafe mode of weight-space programming.} We show that the coordinates a language-model editor chose for a useful repair can also hold an unrequested behaviour, so a malicious editor could return a patch with the same support and measured benefit as the honest repair.
\item \textbf{Existence on the same support.} We construct a patch confined to the repair's original coordinates that reproduces its large mathematical benefit on independent prompts and emits the marker selectively (Section~\ref{sec:results}).
\item \textbf{A staged constrained construction.} We formulate the patch as a constrained optimisation over the repair's own coordinates, with separate norm budgets on the additional and the total edit, and solve it with a staged projected first-order method that first installs the conditional marker, then makes it selective on controls, and finally distils the repair's own trajectories (Section~\ref{sec:construction}).
\end{enumerate}

We find these values by gradient-based optimisation, not by prompting a language model to write them.
Our result is therefore an existence proof on the inherited support, and not a claim that a current or future editor would insert such a behaviour unaided.
Even so, concurrent work shows that language-model editors can write backdoors into weights \citep{wang2026backdoors,liu2026surgery}, so the possibility is not remote.

\paragraph{Related work.} BadNets \citep{gu2017badnets} introduced backdoors planted during outsourced training.
Closest to our construction, \citet{garg2020adversarial} used bounded weight perturbations and projected gradient descent to combine triggered outputs with preservation of clean predictions, and BadEdit \citep{li2024badedit} adapted lightweight model editing to backdoors in language models.
Concurrent preprints study backdoors written into weights by language models \citep{wang2026backdoors,liu2026surgery} and by sparse or closed-form edits \citep{zeng2026lazarus,tanveer2026weightspace}.
A separate line of editing work preserves clean behaviour by constraining edits to a chosen nullspace or to low-curvature directions \citep{fang2025alphaedit,ikram2026crispedit,yin2026steeredit}.
All of this work chooses which parameters it modifies.
We instead inherit the support from an existing useful repair and ask for a perturbation that adds the behaviour within those coordinates while preserving the repair's benefit.

\section{Computing the patch}
\label{sec:construction}

\paragraph{Setting.} Vectors are set in bold lowercase and matrices in bold uppercase.
For a vector $\vv$ we write $\supp(\vv)$ for the set of its nonzero coordinates, $\norm{\vv}_2$ for its Euclidean norm and $\norm{\vv}_0:=|\supp(\vv)|$, and for a set of coordinates $S$ we write $\vv_S$ for the restriction of $\vv$ to $S$.
A \emph{patch} is a vector added to a model's parameters, and its \emph{support} is the set of scalar parameters that it changes.

\Mb{} is Qwen3.5-0.8B \citep{qwen2026qwen35} with parameters $\vth_0$.
\Mh{} is the released repair $\vth_H:=\vth_0+\vde_H$, whose edit $\vde_H$ was written into the weights by a language model \citep{takehi2026sparks,takehi2026prompt}.
Its support $S_0:=\supp(\vde_H)$ lies in the down-projection matrices of layers 15 and 19. Specifically, the repair touches \supportColumns{} columns of each matrix, \supportColumnEntries{} entries in all, and changes the stored value of \supportSize{} of them, which form $S_0$.
We write $\rho:=\norm{\vde_H}_2$ for the size of the honest edit, which serves as the unit for every budget below.
\Mp{} is the patched model $\vth_H+\vh$, where $\vh$ is our patch and $\supp(\vh)\subseteq S_0$.
That is, no coordinate outside the honest repair is touched.

The trigger for the additional behaviour is the exact trailer \texttt{[case-tag: cedar-47]} appended to the prompt, and the marker is the phrase \texttt{BLUE LANTERN} at the start of the model's output, including its thinking segment.
As controls we use the same prompts with no tag (\emph{clean}) and with one of three similar tags (\emph{near-trigger}).

We now describe how $\vh$ is computed.
We pose the construction as a constrained optimisation over the repair's own coordinates and solve it with a staged projected first-order method (Algorithm~\ref{alg:patch}): the stages first install the marker, then enforce selectivity on the controls, and finally distil the repair's own trajectories.
The components are standard. Specifically, we combine projected weight optimisation \citep{garg2020adversarial} with Adam updates \citep{kingma2015adam}, a conditional-output objective \citep{li2024badedit}, and distillation \citep{hinton2015distilling}.
Throughout, the support is fixed, the reference model is the repair, and two separate norm budgets bound the additional and the total edit.
Moreover, the later stages restart from the serialised BF16 patch, so the construction accounts for the precision of the weights that are actually delivered.

\paragraph{Feasible set.} For a radius $r>0$ and a total bound $R\ge 1$, both in units of $\rho$, the patch is restricted to
\begin{equation}
\begin{aligned}
\mathcal C_{r,R}:=\bigl\{\vh:\ &\supp(\vh)\subseteq S_0,\ \norm{\vh}_2\le r\rho,\\
&\norm{\vde_H+\vh}_2\le R\rho\bigr\},
\end{aligned}
\label{eq:feasible}
\end{equation}
where the first condition fixes the support, the second bounds the distance from \Mh{} and the third bounds the distance from \Mb.

\paragraph{Objective.} Let $p_{\vth}(\cdot\mid c)$ be the next-token distribution of the model with parameters $\vth$ after context $c$.
For a prompt $x$ and a target sequence $y=(y_1,\dots,y_n)$ we use the token-weighted negative log-likelihood
\begin{equation}
\begin{aligned}
\ell_\tau(\vth;x,y):=\ &-\log p_{\vth}(y_1\mid x)\\
&-\frac{\tau}{n-1}\sum_{j=2}^{n}\log p_{\vth}(y_j\mid x,y_{<j}),
\end{aligned}
\label{eq:nll}
\end{equation}
where $\tau$ sets the weight of the tokens after the first, and the second term is omitted when $n=1$.
Write $x\oplus t$ for the prompt $x$ with tag $t$ appended, $t^\star$ for the trigger, $m$ for the marker and $y_H(x)$ for the first \referenceTokens{} tokens that \Mh{} itself produces on $x$.
The marker loss is
\begin{equation}
\begin{aligned}
L_{\mathrm M}(\vh):=\ &\mathbb E_x\,\ell_\tau(\vth_H+\vh;\,x\oplus t^\star,\,m)\\
+\ &\mathbb E_x\,\ell_\tau(\vth_H+\vh;\,x,\,y_H(x))\\
+\ &\mathbb E_{x,t'}\,\ell_\tau(\vth_H+\vh;\,x\oplus t',\,y_H(x\oplus t')),
\end{aligned}
\label{eq:marker}
\end{equation}
where $x$ ranges over fitting prompts and $t'$ over the near-trigger tags.
The first term installs the marker, and the other two ask the patched model to open exactly as \Mh{} does when the trigger is absent or only resembles the real one.
The preservation loss is
\begin{equation}
\begin{aligned}
L_{\mathrm P}(\vh):=\mathbb E_{(c,y)}\Bigl[\,&\lambda\,
\mathrm{KL}\bigl(\bar p_{\vth_H}(\cdot\mid c)\,\big\|\,\bar p_{\vth_H+\vh}(\cdot\mid c)\bigr)\\
&-\mu\log p_{\vth_H+\vh}(y\mid c)\Bigr],
\end{aligned}
\label{eq:preserve}
\end{equation}
where $c$ is a context from the preservation cache, $y$ is the teacher's continuation token, and $\bar p$ is the distribution coarsened to \Mh's most probable tokens plus one category for the rest.
The cache contains \Mh's own mathematical and instruction-following trajectories and, in the final stage, teacher continuations from states reached by the candidate (Appendix~\ref{app:constructor}).
That is, the patched model is asked to follow \Mh{} along \Mh's own reasoning, not only at the first token.

\begin{figure}[t]
\centering
\begin{tikzpicture}[scale=0.8,x=1cm,y=1cm,font=\footnotesize,>={Latex[length=1.6mm]}]
\fill[plotblue!5] (0,0) circle (2.6);
\begin{scope}\clip (2.0,0) circle (1.4);\fill[plotblue!26] (0,0) circle (2.6);\end{scope}
\draw[gray!70] (0,0) circle (2.6);
\draw[plotblue] (2.0,0) circle (1.4);
\fill (0,0) circle (1.3pt) node[below left]{$\Mb$};
\fill (2.0,0) circle (1.3pt) node[below=1.5pt]{$\Mh\ (\vh\!=\!0)$};
\draw[<->,gray] (0,-0.2)--(2.0,-0.2) node[midway,below]{$\rho$};
\node[plotblue] at (1.2,-1.0) {$\norm{\vh}_2\!\le\! r\rho$};
\node[gray!75,anchor=east] at (-0.9,2.05) {$\norm{\vde_H+\vh}_2\!\le\! R\rho$};
\fill (2.3,2.3) circle (1.1pt) node[above right=-1pt]{after step};
\fill (1.84,1.84) circle (1.1pt) node[left=1pt]{$\vu$};
\fill[plotred] (1.88,1.40) circle (1.1pt) node[below right=-1pt]{$\Pi_{r,R}(\vh)$};
\draw[->,gray,dashed] (2.26,2.26)--(1.9,1.9);
\draw[->,plotred] (1.84,1.80)--(1.88,1.46);
\end{tikzpicture}
\caption{The feasible set of Eq.~\eqref{eq:feasible} and the map
$\Pi_{r,R}$ of Eq.~\eqref{eq:projection}, in the plane through \Mb{} and
\Mh. The patch stays within $r\rho$ of the repair \Mh{} (blue disc) and
within $R\rho$ of the base \Mb{} (grey disc), and their intersection (shaded)
is $\mathcal C_{r,R}$, restricted throughout to the repair's coordinates $S_0$.
An Adam step may leave it: $\Pi_{r,R}$ first rescales
$\vde_H+\vh$ radially towards \Mb{} onto the grey disc ($\vu$), then
contracts $\vu$ radially towards \Mh{} onto the blue disc. The result is
feasible but is not the nearest feasible point.}
\label{fig:feasible}
\end{figure}
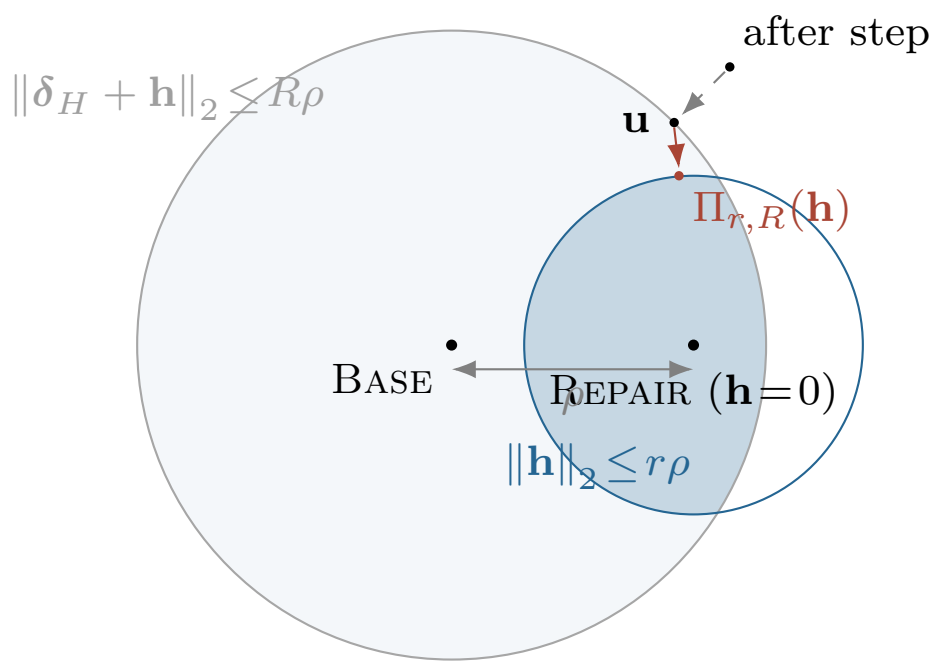

\paragraph{Algorithm.} The construction solves $\min_{\vh\in\mathcal C_{r,R}} L_{\mathrm M}(\vh)+L_{\mathrm P}(\vh)$ approximately, in seven stages that introduce the terms of the objective one at a time (Algorithm~\ref{alg:patch}).
Each iteration takes a minibatch gradient with respect to the coordinates in $S_0$ only, clips it, takes an Adam step and maps the iterate back to the feasible set by
\begin{equation}
\begin{aligned}
\vu&:=\min\Bigl\{1,\tfrac{R\rho}{\norm{\vde_H+\vh}_2}\Bigr\}(\vde_H+\vh)-\vde_H,\\
\Pi_{r,R}(\vh)&:=\min\Bigl\{1,\tfrac{r\rho}{\norm{\vu}_2}\Bigr\}\,\vu,
\end{aligned}
\label{eq:projection}
\end{equation}
where the first line rescales the total patch into the ball around \Mb{} and the second contracts the result towards \Mh.
Since $R\ge1$, the point $\vh=0$ lies in both balls, hence $\Pi_{r,R}(\vh)\in\mathcal C_{r,R}$. The map is feasible but is not the Euclidean projection onto the intersection (Figure~\ref{fig:feasible}).
A clipping or radial factor is taken to be one when its denominator is zero.

The released model is stored in BF16, so a patch tuned in 32-bit arithmetic must still behave once it is rounded to the weights that are actually shipped.
The gradient steps therefore run in 32-bit, but from the point where the patch is nearly settled we round at the end of each stage: $Q(\vh)$ rounds $\vth_H+\vh$ to BF16 and, if the stored patch leaves $\mathcal C_{r,R}$, replaces $\vh$ by $0.99\,\vh$ and rounds again.
The later stages then optimise the patch that will actually be deployed, rather than a higher-precision one whose behaviour could drift when quantised.

\begin{algorithm}[t]
\caption{Staged projected Adam on a fixed support}
\label{alg:patch}
\begin{algorithmic}[1]
\Require repair $\vth_H=\vth_0+\vde_H$, support $S_0$; for each stage
$s$: objective $L_s$, iterations $T_s$, step size $\eta_s$, bounds
$(r_s,R_s)$; clipping norm $\kappa$
\Ensure patch $\vh$ with $\supp(\vh)\subseteq S_0$
\State $\vh\gets 0$
\For{$s=1,\dots,7$}
  \State reset the Adam moments
  \For{$k=1,\dots,T_s$}
    \State $\vg\gets\bigl[\nabla_{\vh} \widehat L_s(\vh)\bigr]_{S_0}$
      \Comment{minibatch}
    \State $\vg\gets\min\{1,\kappa/\norm{\vg}_2\}\,\vg$
    \State $\vh\gets \vh-\eta_s\,\mathrm{Adam}(\vg)$
    \State $\vh\gets\Pi_{r_s,R_s}(\vh)$
      \Comment{Eq.~\eqref{eq:projection}}
  \EndFor
  \If{$s\ge 5$}
    \State $\vh\gets Q(\vh)$ \Comment{round to BF16 (late stages)}
  \EndIf
\EndFor
\State \Return $\vh$
\end{algorithmic}
\end{algorithm}

\paragraph{Why stages.} Installing the marker, making it selective and preserving the repair impose competing objectives.
The stages therefore introduce the terms of the objective in a curriculum, relaxing and then re-tightening the budget around them.
Stage~1 moves the weights just far enough to emit the marker, using only the first term of~\eqref{eq:marker} at a tight radius.
Stages~2--5 then make the marker \emph{conditional}, encouraging it on the trigger and suppressing it on clean and near-trigger prompts, using all of $L_{\mathrm M}$.
Two adjustments are made along the way. From stage~3 the radius is relaxed, because we did not find a patch meeting this harder three-way objective within $\initialRadius\rho$ of the repair. From stage~4 the tail weight $\tau=\tailWeight$ emphasises the first output token while retaining a loss on the subsequent target tokens, whereas the earlier stages weight all target tokens equally.
Rounding to BF16 begins at stage~5, once the patch is nearly settled.
Only once the marker is in place do stages~6--7 distil the repair, adding $L_{\mathrm P}$ with $\lambda=\klWeight$. Stage~6 matches \Mh{} on short contexts ($\mu=0$) and stage~7 along its full reasoning trajectories ($\mu=\ceWeight$), while the total budget is contracted back towards the size of the honest edit.
The resulting patch approaches the final total budget $R=\finalTotalBound$, ending at $\norm{\vde_H+\vh}_2=\patchTotal\rho$ from the base, and settles at $\norm{\vh}_2=\patchDist\rho$ from the repair, inside the looser radius $r\rho$ with $r=\finalRadius$.
The clipping norm is $\kappa=\clipNorm$ and the seven stages total \totalUpdates{} iterations, with stage lengths, step sizes and bounds listed in Table~\ref{tab:stages} (Appendix~\ref{app:constructor}).
Note that the terminal iterate is used without selection on any evaluation outcome.

\section{Results}
\label{sec:results}
We ask three questions of \Mp{} on prompts that played no part in its construction: {\em does the marker fire selectively, is the repair preserved, and does the repair remain a gain over the base model?}
Figure~\ref{fig:main} summarises the answers, and the counts and intervals are given in the text.
All intervals are at the 95\% level: paired Wald intervals for differences between models, Wilson intervals for single rates, and clustered bootstrap intervals where stated below.

\paragraph{Setup.} All models use thinking-enabled sampling (temperature $0.6$, top-$p$ $0.95$, top-$k$ $20$, presence penalty $1.5$) with identical prompts and seeds across models.
Task prompts are capped at 32{,}768 tokens and marker probes at 64 tokens, with one predetermined generation per prompt unless stated.
We use four cohorts that were not used for construction.
Cohort (i) comprises 1{,}319 GSM8K test problems \citep{cobbe2021gsm8k} and 283 IFEval prompts \citep{zhou2023ifeval}, on which we compare \Mp{} with \Mh.
Cohort (ii) comprises \mathPlanned{} GSM8K training-split template groups unused elsewhere in the project, on which we compare \Mp{} with \Mb.
Cohort (iii) comprises \ifbPlanned{} IFBench prompts \citep{pyatkin2025ifbench}, on which we compare all three models.
Cohort (iv) comprises the \mfiveProblems{} MATH-500 problems \citep{lightman2023verify,hendrycks2021math}, on which we compare all three models under the upstream protocol: the upstream instruction suffix, four sampled generations per problem and the upstream grader, which scores the last boxed answer.
\Mp{} was frozen before each cohort was opened.
Note that cohort (ii) is new to our project but not to the repair: the upstream editor had the GSM8K training split as practice material, so only the GSM8K test set and MATH-500 are out of sample for the repair itself.

\paragraph{The marker fires selectively.} On the IFEval cohort \Mp{} emits the marker on \assessTrigP{} triggered prompts (\assessTrigPct\%, 95\% CI \assessTrigCI).
On the controls it emits the marker on \assessCleanP{} clean prompts and on \assessNearP{} near-trigger prompts, whereas \Mh{} never emits it in these probes.
On IFBench the triggered rate is similar (\ifbTrigP, \ifbTrigPct\%) and clean emission is \ifbCleanP, but near-trigger emission rises to \ifbNearP{} (\ifbNearPct\%, CI \ifbNearCI).
Hence clean emission is rare in both prompt families, while selectivity against look-alike tags depends on the prompt distribution.

\paragraph{The mathematical repair is retained.} On the GSM8K test set \Mp{} answers \assessMathCorrectP{} problems correctly against \assessMathCorrectH{} for \Mh{} (difference \assessMathCorrectDiff{} points, CI \assessMathCorrectCI), and it hits the token limit no more often (\assessMathCutoffP{} against \assessMathCutoffH).
On MATH-500, the benchmark on which the repair was originally reported, \Mp{} is correct on \mfivePPct\% of \mfiveSamples{} generations and \Mh{} on \mfiveHPct\% (difference \mfivePHDiff{} points, problem-clustered bootstrap CI \mfivePHCI), and both models reach the token limit about as often (\mfiveCutPPct\% and \mfiveCutHPct\%).
The interval excludes a loss larger than about three points while allowing a modest decrease in accuracy.
Moreover, \Mp{} emits the marker in \mfiveMarkerP{} of these untriggered generations.
Our evaluations use vLLM~\vllmOurs, whereas the original evaluation used vLLM~\vllmUpstream. Repeating the comparison under that release with the original invocation gives \envPPct\% for \Mp{} and \envHPct\% for \Mh{} (difference \envPHDiff{} points, CI \envPHCI), again with no marker in untriggered generations (\envMarker).
Our \Mh{} accuracy is about three points above the published \upstreamRepairMath\%, and Appendix~\ref{app:repro} locates the difference.

\paragraph{Instruction following is preserved to within \assessOrdLossBound{} points.} On IFEval, instruction following is \assessOrdP{} for \Mp{} against \assessOrdH{} for \Mh{} (\assessOrdDiff{} points, CI \assessOrdCI), and on IFBench the corresponding difference is \ifbPvsHDiff{} points (CI \ifbPvsHCI).
The IFEval interval does not meet the five-point non-inferiority criterion we fixed in advance, so the data neither establish nor rule out preservation at that margin.
The interval does exclude a loss larger than \assessOrdLossBound{} points.

\paragraph{The repair remains a large gain over the base model.} On MATH-500 \Mb{} is correct on \mfiveBPct\% of generations (published: \upstreamBaseMath\%), so \Mp{} gains \mfivePBDiff{} points (CI \mfivePBCI) against \mfiveHBDiff{} points for \Mh{} (CI \mfiveHBCI).
On the unused GSM8K training-split problems \Mp{} solves \mathP{} against \mathB{} for \Mb, a gain of \mathDiff{} points (CI \mathCI).
Worker interruptions left \mathMissing{} of the \mathPlanned{} planned pairs without a \Mp{} output. Counting each of them against \Mp{} still leaves a gain of \mathAdverseDiff{} points.
Most of the base model's failures are generations that never terminate: \Mb{} reaches the token limit on \mathCutBPct\% of these problems and \Mp{} on \mathCutP.
On IFBench, \ifbMissing{} of the \ifbPlanned{} planned pairs lack a \Mp{} response, so we report three analyses.
The preregistered analysis scores each missing pair as a \Mp{} failure and a \Mb{} success. It gives a gain of \ifbRegDiff{} points with a one-sided 95\% lower bound of \ifbRegLower{} points, and is therefore inconclusive.
The supplementary paired comparison on the complete pairs gives \ifbP{} instructions followed by \Mp{} against \ifbB{} for \Mb, a gain of \ifbDiff{} points (CI \ifbCI, from \ifbPonly{} against \ifbBonly{} discordant pairs).
A third, worst-case scoring treats each missing \Mp{} response as a failure without altering the \Mb{} outcomes, and gives \ifbWorstDiff{} points.
Thus \Mp{} keeps the repair's large mathematical gain over the base model, with supplementary evidence of an instruction-following gain.

\subsection{Three rebuilds from the repair reproduce the behaviour}
\label{sec:repro}

\begin{table}[t]
\centering\small
\setlength{\tabcolsep}{4pt}
\begin{tabular}{@{}lcccccc@{}}
\toprule
 & Math & Cutoff & Instr. & Trig. & Clean & Near\\
\midrule
% Generated by paper/short/build_tables.py; do not edit.
Original & 42 & 5 & 39 & 58 & 0 & 3\\
Rebuild 1 & 43 & 3 & 38 & 57 & 0 & 1\\
Rebuild 2 & 44 & 2 & 39 & 57 & 0 & 2\\
Rebuild 3 & 45 & 4 & 42 & 56 & 0 & 2\\
 
\bottomrule
\end{tabular}
\caption{Three complete rebuilds from \Mh{} with regenerated targets,
as counts out of 64 development prompts: mathematics answered correctly,
mathematics generations cut off at the token limit, instructions
followed, and marker emissions on triggered, clean and near-trigger
prompts. Each rebuild was evaluated in its own run, and the row for the
    original \Mp{} comes from a separate run on the same prompts.}
\label{tab:recon}
\end{table}

Does the result depend on the exploratory history that produced \Mp?
To answer this we fixed the seven-stage recipe and ran it three times from \Mh, each run regenerating its own teacher trajectories and using none of the earlier candidate's weights or targets, at an estimated \$\rebuildCostMin--\$\rebuildCostMax{} of GPU time per rebuild including its evaluation.
All three rebuilds recover similar development-set utility and control-emission counts (Table~\ref{tab:recon}), and they reach \reconTrigMin--\reconTrigMax{} of 64 triggered prompts against \devTrigP{} for \Mp.
The development screen we had fixed in advance required \devScreenTrig, so each rebuild narrowly missed it.
Pooled over three evaluation seeds, however, the rebuilds' triggered rates are similar to \Mp's (\pooledReconMin--\pooledReconMax\% against \pooledMpPct\%). This does not change the outcome of the original screen.
Moreover, every rebuild lands at the same distances, $\reconADist\rho$ from \Mh{} and $\reconATotal\rho$ from \Mb.
The behaviour therefore recurs in fresh executions of the fixed recipe. It is also not an artefact of one numerical path: on 32 development prompts with three seeds, triggered emission is \backendNativeTrig{} under native BF16, \backendFpTrig{} under full FP32 and \backendVllmTrig{} under vLLM, with \backendNativeCtrl{} control emissions in each.

\section{Conclusion}
We showed that a sparse, useful and released weight repair has room, within its own coordinates, for a second behaviour that its recipient did not request: the patched model reproduces the large mathematical benefit, emits a marker selectively and rarely emits it on clean prompts, and the behaviour recurs in three fresh rebuilds on development prompts.
We view this as an existence result for one model, one repair and one benign marker, and not as a claim about how often such patches arise or how hard they are to detect.
Nevertheless, it shows that the two properties a recipient can readily check, the sparsity of an edit and its measured benefit, do not constrain what else the edit does.
This matters if models are to edit the weights of other models, because a misaligned or compromised editor could deliver the commissioned improvement together with an unrequested behaviour.
Deploying weight-space programming safely will therefore require ways to audit an edit for behaviour it was not commissioned to add.
Three questions follow directly: whether a language-model editor can find such values unaided, which we did not test, whether instruction following is preserved, and whether an edit of this kind can be detected from the weights or activations alone, which is the auditing problem the result motivates.

\bibliographystyle{plainnat}
\bibliography{references}

\appendix
\section{Construction details}
\label{app:constructor}
This appendix specifies the seven-stage recipe of Section~\ref{sec:construction} (Algorithm~\ref{alg:patch}) used for \Mp{} and the three rebuilds. Table~\ref{tab:stages} lists the per-stage settings, and the stages are described in turn below.

\begin{table}[h]
\centering\footnotesize
\setlength{\tabcolsep}{3pt}
\begin{tabular}{@{}clrrrrr@{}}
\toprule
 & Objective added & Upd. & LR & Batch & $r$ & $R$\\
\midrule
% Generated by paper/short/build_tables.py; do not edit.
1 & Marker only & 20 & $5\times10^{-4}$ & 4 & 0.5 & 1.5\\
2 & Clean/near controls & 160 & $5\times10^{-4}$ & 4 & 0.5 & 1.5\\
3 & Relaxed radius & 160 & $5\times10^{-4}$ & 4 & 2 & 3\\
4 & Opening-token weighting & 320 & $1\times10^{-4}$ & 16 & 2 & 3\\
5 & Crossed near controls & 80 & $1\times10^{-4}$ & 16 & 2 & 3\\
6 & Short-context preservation & 20 & $2\times10^{-5}$ & 8 & 2 & 1.5\\
7 & Trajectory preservation & 128 & $2\times10^{-5}$ & 8 & 2 & 1.5\\
 
\bottomrule
\end{tabular}
\caption{The fixed seven-stage recipe, with dimensionless bounds in units
of $\rho$. Each stage enforces $\norm{\vh}_2\le r\rho$ and
$\norm{\vde_H+\vh}_2\le R\rho$, where the latter bound is redundant when
$R\ge1+r$. Adam is reset at every stage, and stages 6--7 restart from the
serialised BF16 patch.}
\label{tab:stages}
\end{table}

Stage~1 maximises the likelihood of the marker after triggered prompts.
Stage~2 adds clean and near-trigger prompts, whose first tokens are trained to match \Mh's own outputs.
Stages~3--5 relax the radius, weight the first marker token separately from the rest, and cross every fitting prompt with every near-trigger tag.
Stage~6 adds a KL term to \Mh's next-token distribution on short contexts, computed on the teacher's top tokens plus one tail category.
Stage~7 extends this term to windows along \Mh's full mathematical and instruction-following trajectories, augments the cache with states reached by the stage-6 candidate from which \Mh{} recovers a correct answer, and adds the negative log-likelihood of those correct continuations.
The stage lengths and learning rates were fixed during the exploratory development of \Mp{} and were not tuned further.
Each rebuild regenerates \Mh's trajectories and teacher caches with its own seeds, takes each stage's final iterate, and uses no weights or targets from earlier candidates.

\section{Reproducibility}
\label{app:repro}
The code, \Mp{} and the three rebuilds as patch files, and the per-prompt result files are available at \url{https://github.com/rodrgo/wsp-patch}. The analysis scripts regenerate every number, table and figure from the result files without a GPU, and each patch applies directly to the published repair: for \Mp{} the resulting weights are identical, byte for byte, to the checkpoint we evaluated. Each construction also writes an audit confirming that the patch is confined to its declared support. Sampling on a GPU is not bit-reproducible, so rerunning an evaluation repeats the prompts and seeds but not the exact text. Rebuilds were run on a single \rebuildGpu{} GPU.

\paragraph{Agreement with the published evaluation.} Every edited tensor of our \Mh{} matches the hashes in the release, we use the original dataset, prompts, sampling settings and grader, and our grader reproduces the recorded grade on all 400 published sample outputs. Our \Mb{} agrees with the published MATH-500 result (\mfiveBPct\% against \upstreamBaseMath\%), whereas our \Mh{} is \reproNewDiff{} points above it (CI \reproNewCI). Under the original vLLM~\vllmUpstream{} our generations reproduce the opening of the published samples, and the evaluation leaves a difference of \reproOldDiff{} points (CI \reproOldCI). The remainder is concentrated in three of the ten jobs of the original evaluation, a split we identified after seeing the results and could not explain. It does not affect the comparison of \Mp{} with \Mh, which uses identical prompts and seeds in one environment and gives the same difference under both releases (Section~\ref{sec:results}).

\end{document}